\documentclass[10pt,twocolumn,letterpaper]{article}

\usepackage{cvpr}
\usepackage{times}
\usepackage{epsfig}
\usepackage{graphicx}
\usepackage{subcaption}
\usepackage{amsmath}
\usepackage{dsfont}
\usepackage{amssymb}
\usepackage{mathrsfs}
\usepackage{tabularx}
\usepackage{makecell}
\usepackage{bm}
\usepackage[dvipsnames]{xcolor}

\usepackage{caption}

\mathchardef\mhyphen="2D 

\usepackage[pagebackref=true,breaklinks=true,letterpaper=true,colorlinks,bookmarks=false]{hyperref}

\cvprfinalcopy 

\def\cvprPaperID{2552} 

\ifcvprfinal\fi
\begin{document}

\title{Learning to Forecast Videos of Human Activity with Multi-granularity Models and Adaptive Rendering}


\author{Mengyao Zhai, Jiacheng Chen, Ruizhi Deng, Lei Chen, Ligeng Zhu, Greg Mori\\
Simon Fraser University\\
}

\definecolor{Mycolor2}{HTML}{01F9DE}
\newcommand{\gm}[1]{\textcolor{blue} {#1}}
\newcommand{\mz}[1]{\textcolor{magenta} {#1}}
\newcommand{\lc}[1]{\textcolor{cyan} {#1}}
\newcommand{\jc}[1]{\textcolor{red} {#1}}
\newcommand{\rz}[1]{\textcolor{green} {#1}}

\maketitle

\begin{abstract}
We propose an approach for forecasting video of complex human activity involving multiple people. Direct pixel-level prediction is too simple to handle the appearance variability in complex activities. Hence, we develop novel intermediate representations. An architecture combining a hierarchical temporal model for predicting human poses and encoder-decoder convolutional neural networks for rendering target appearances is proposed. Our hierarchical model captures interactions among people by adopting a dynamic group-based interaction mechanism. Next, our appearance rendering network encodes the targets' appearances by learning adaptive appearance filters using a fully convolutional network.  Finally, these filters are placed in encoder-decoder neural networks to complete the rendering. We demonstrate that our model can generate videos that are superior to state-of-the-art methods, and can handle complex human activity scenarios in video forecasting.
\end{abstract}





\begin{figure}[h]
\center
\includegraphics[scale=0.4]{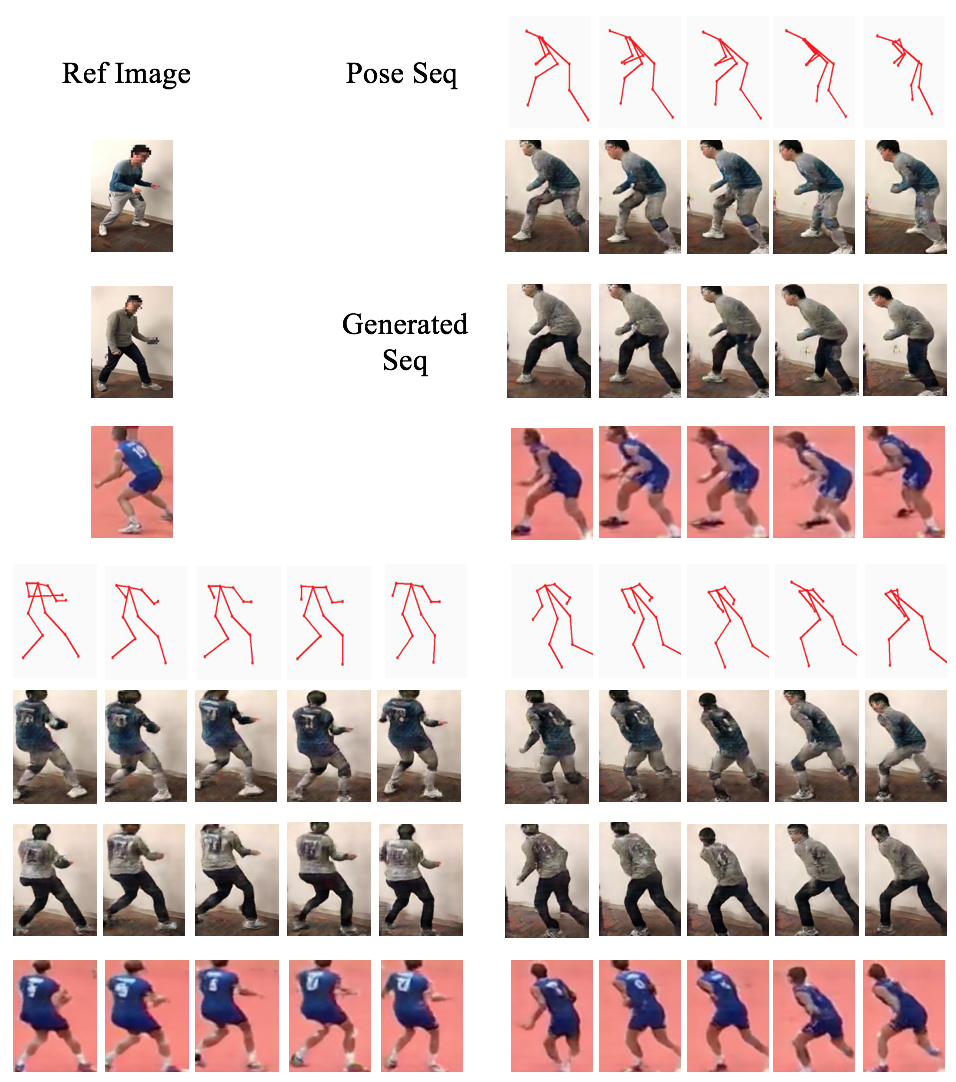}
\caption{We develop architectures for forecasting videos of complex human activity.  At the core of the method is adaptive rendering modules.  Given a reference image and a forecasted pose sequence, we can generate realistic video forecasts of the reference person.  A series of examples of various reference images rendered into novel pose sequences is shown above.}
\label{fig-intro}
\end{figure}


\section{Introduction}

We may not be able to play soccer like Lionel Messi, but perhaps we can train deep networks to hallucinate imagery suggesting that we can.  Consider the images in Fig.~\ref{fig-intro}.  In this paper we describe research toward synthesizing realistic sequences that forecast the appearance of people performing complex actions.  The model can predict the future poses of people, and use sample appearance images to generate novel views of people that can capture fine details such as imaginary numbers that appear on the backs of people's clothing.

Future prediction is a fundamental and important problem in many domains. Determining what will happen next can enable myriad applications.  Recent examples of attempts to model this predictive process exist across a variety of research fields.  Within robotics, work has explored predicting the consequences after interactions between an agent and its environment~\cite{finn2016unsupervisedPhysicalInteraction}. In natural language processing, approaches~\cite{mansimov2015generating,reed2016generative} have been proposed to tackle tasks such as text to image or image to text synthesis. Accurate generative models of video sequences are a core part of visual understanding and have received renewed attention from the vision community~\cite{walker2017poseKnow, Villegas2017LearningToGenerate}.

In this paper, we focus on learning how to forecast videos of human actions in complex scenarios.  Sports videos are an ideal setting for this study: complex in terms of multiple targets, rich in interactions, motion blur, and appearance variation.  How to understand the patterns presented in sports videos and provide cues for the prediction of subsequent frames are of key importance here.  Moreover, developing generation models that can realize the substantial variability in image content that arise from human body articulation and appearance variation is a challenge.

We address these challenges by developing a video forecasting approach with two main novel components.  Human body pose is a natural intermediate representation for this forecasting, and hence utilized in many previous methods for synthesizing human motion and video~\cite{BrandH00,EfrosBMM03,walker2017poseKnow}.  We follow in this paradigm, predicting body poses and using them to generate video sequences of future human motion.

First, since we address complex video forecasting, we develop a novel hierarchical recurrent neural network structure that can model multiple people as well as their interactions. This structure captures levels of detail ranging from group-level dynamics down to predictions on individual human body joints.  The first layer of our model captures group inference and predicts future poses by leveraging an interaction context.
We devise a dynamic group-based interaction mechanism where people dynamically change groups according to the likelihood of interacting with people in that group, and the likelihood is estimated using both pose and location information.
The second layer is a structured spatio-temporal LSTM~\cite{liu2016st-LSTM}, predicting finer adjustments for first-layer results to refine the prediction for each human joint.



After pose prediction, the core task is to generate realistic images of a particular person striking this pose. Simple networks~\cite{walker2017poseKnow} may generate blurry and distorted images.  Stylistic methods~\cite{isola2016pix2pix} have shown great success in generating realistic images, but lack control over the appearance of the generated images. Our task requires the model to be able to generate images of a person with a specific appearance. Inspired by~\cite{chen2017stylebank}, we propose a novel appearance rendering network which encodes appearance into convolutional filters. These filters are operationalized using a fully convolutional network, and utilized in an image-to-image translation structure that transfers the desired appearance to the generated image.

To sum up, we contribute a new state of the art generative model that (1) focuses on forecasting videos of complex human activities involving multiple people; (2) dynamically infers group memberships; and (3) performs adaptive appearance rendering to create accurate depictions of human figures in these forecasted poses.

\begin{figure*}[!tph]
\center
\includegraphics[width=0.9\textwidth]{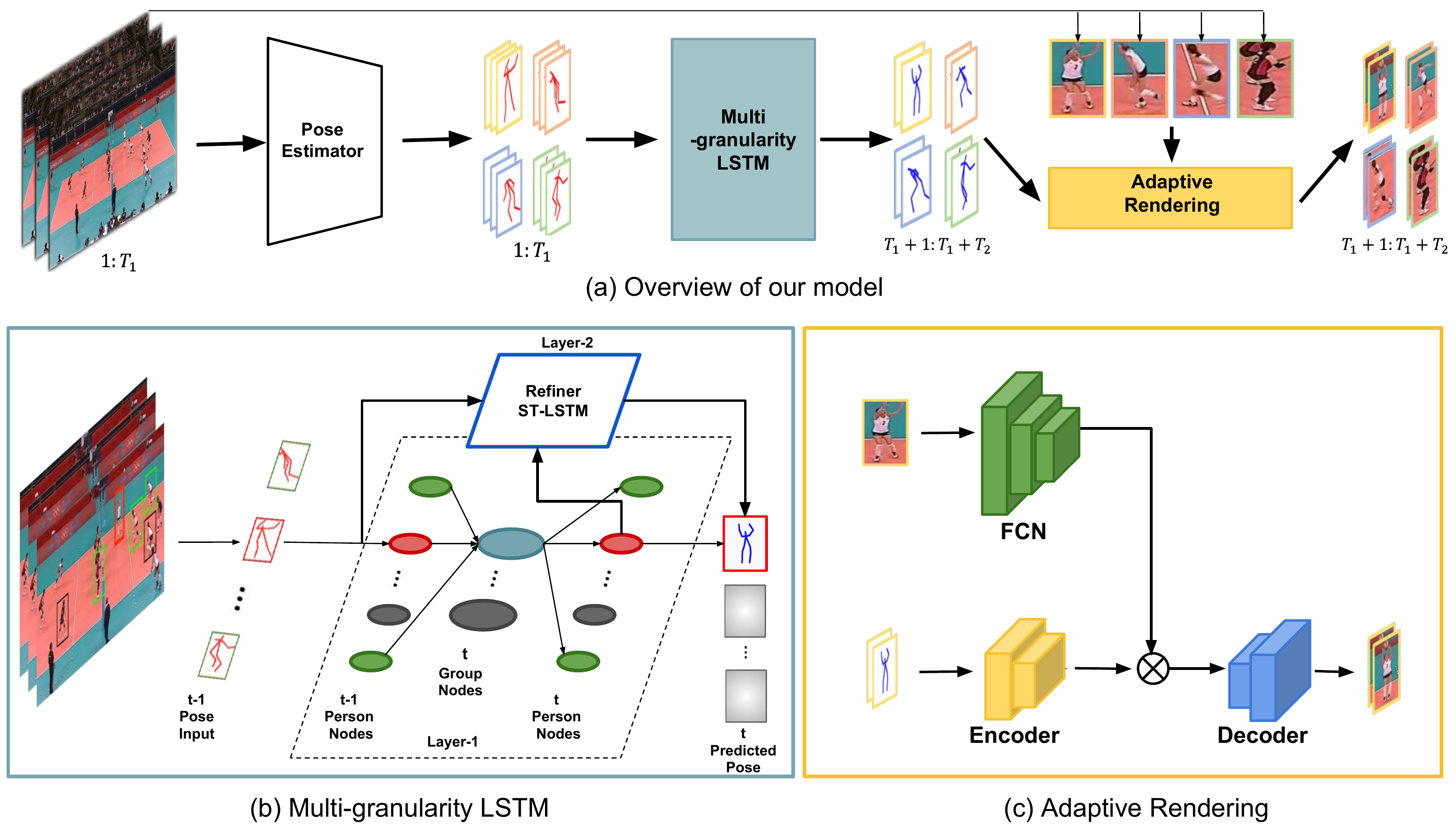}
\caption{(a) Overall hierarchical model for human activity video forecasting. Given input frames, the poses of each person in the scene are estimated, then (b) our multi-granularity LSTM predicts future poses of each person (temporal links for LSTM nodes are omitted. \textbf{\textcolor{red}{red}} node denotes the person currently being predicted, and \textbf{\textcolor{green}{green}} nodes denote people in the same group with red while \textbf{\textcolor{black}{black}} nodes denote people not in the same group; red and green nodes are connected to the \textbf{\textcolor{blue}{blue}} group node representing their group), followed by (c) our adaptive rendering network, generating realistic images of each person striking the predicted pose.}
\label{fig:overview}
\end{figure*}



\section{Related Work}

{\bf Video forecasting:} Data-driven video prediction has seen a renaissance in recent years. One major branch of methods uses RNN-based models such as encoder-decoder LSTMs for direct pixel-level video prediction~\cite{ranzato2014videoGenerationBaseline, srivastava2015unsupervisedVideoRepresentationLSTMs, oh2015actionConditionedVideoPrediction, mathieu2015deepBeyondMSE}. Another type of approach~\cite{xue2016VisualDynamics} models future frames in a probabilistic manner. These methods successfully synthesized low-resolution videos with relatively simple semantics, such as moving MNIST digits or human action videos with very regular, smooth motion.

Subsequent work has attempted to expand the quality of predicted video in terms of resolution and diversity in human activity.  Earlier efforts were focused on optical flow-timescale prediction, further work pushed past into more complex motions (e.g.~\cite{walker2016uncertainFuture, liu2017video}).

Predicting video frames directly in low-level pixel space is difficult and these types of approaches tend to generate blurry or distorted future frames. To tackle this problem, hierarchical models~\cite{walker2017poseKnow, Villegas2017LearningToGenerate} adopt intermediate representations. These models generate future frames in two stages: first, future poses are generated, then binary pose images are transformed into realistic frames. This type of approach can alleviate image blur, however the quality of generation largely depends on the the image generation network. Simple generation networks can still produce blurry images as shown in~\cite{walker2017poseKnow}. 

Further difficulties arise in generating accurate human poses.  Previous generative approaches use simplistic temporal pose models.  Within the field of 3D action recognition, human pose sequences are subjected to spatio-temporal analysis~\cite{liu2016st-LSTM, liu2017global}. Specifically, structure-based spatio-temporal LSTMs are effective for robust processing of human body joint position data~\cite{liu2016st-LSTM}. 

{\bf Modeling human interactions:} In this paper, we propose an architecture for predicting the future of multi-person video. We introduce a novel human-human interaction mechanism as well as a flexible image-to-image translation model. Previous work on human-human interactions includes the SocialLSTM~\cite{Alahi2016Social}, a generic data-driven approach for modeling interaction among pedestrians. Jain et al.~\cite{jain2016structural} proposed a rich RNN mixture which is a spatio-temporal graph for modeling object-object interactions across time. Deng et al.~\cite{deng2016structure} proposed a structure learning model where pair-wise interactions are learned and relations among persons are determined by imposing gates.

{\bf Generative image models:} Image-to-image translation has achieved great success since the emergence of GANs~\cite{goodfellow2014GAN}. Recent work produces promising results using GAN-based models~\cite{isola2016pix2pix, CycleGAN2017}. Stylized images can be generated by using feed-forward networks~\cite{gatys2015neuralStyleTransfer} with the help of perceptual loss~\cite{johnson2016perceptualLoss}. The recent work of \cite{chen2017stylebank} proposes a structure to disentangle style and content for style transfer. Styles are encoded using a stylebank (set of convolution filters). Visual analogy making~\cite{NIPS2015_5845_analogy, sadeghi2015visalogy} generates or searches for an new image analagous to an input one, based on other previously given example pairs.

In summary, our approach builds on the substantial body of related work in pose analysis, group interaction, and style/analogy-based image generation. We contribute a hierarchical method for pose prediction from the person-interaction down to the body joint level, combined with a novel adaptive appearance rendering model for image generation.

\section{Forecasting Complex Human Activity}

We propose a method for generating videos of complex human activities.  An overview of the method is shown in Fig.~\ref{fig:overview}.  The input to our method is a video sequence of multiple people.  Human poses are obtained using state-of-the-art techniques.  From there, we first forecast poses with our multi-granularity model (Sec.~\ref{sec:multigran}).  The predicted poses are rendered into images with our adaptive rendering technique (Sec.~\ref{sec:adaptive_render}).  This image synthesis technique is general, and can be utilized in other paradigms (e.g.\ inserting novel people, appearance adaptation), which are explored in our experiments in Sec.~\ref{sec:experiments}.

\subsection{Multi-granularity Pose Prediction}
\label{sec:multigran}
%
%
We propose a multi-granularity model for predicting future pose for multiple targets.  This is a hierarchical model that reasons over
groups of people and uses this to predict future poses. The predictive process is illustrated in Fig~\ref{fig:overview}(b). The first layer is equipped with a group-based dynamic interaction mechanism for modeling inter-person interactions. The second, intra-person layer is a refiner spatio-temporal LSTM that refines the generation from the first layer. 
%

\subsubsection{Group Interaction Mechanism}
\label{sec:multigran-group-mechanism}

\begin{figure}[h]
\center
\includegraphics[width=\linewidth, height=0.55\linewidth]{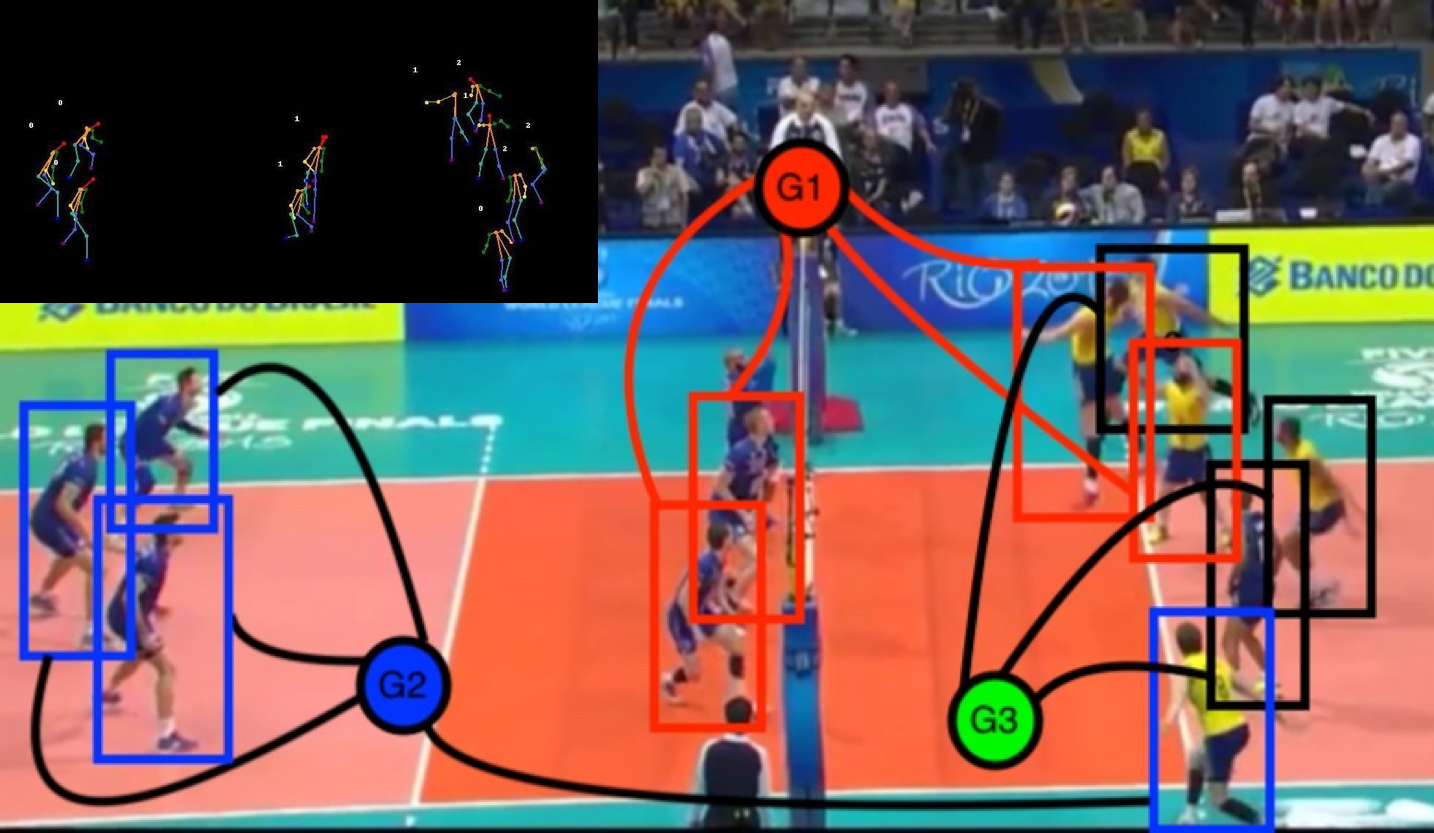}
\caption{Visualization of the group-based interaction mechanism.}
\label{fig-group}
\end{figure}

For complex human activity, analyzing relations among people can be beneficial. Given a set of people in a scene, not all people in the scene are interacting with each other, hence a mechanism for automatically inferring relations is important. As shown in Fig.~\ref{fig-group}, which is produced by our group-based interaction mechanism, our model learns to assign all people into groups. People having strong interactions with each other are learned to be grouped together so that information aggregated over each group can help better predict future poses for its members.

Given $N$ people in a scene, we define $N-1$ groups for representing potential interactions. We use $G_{k}^t$ to denote the $k^{th}$ group at time $t$ and $||G_{k}^{t}||$ is the size of the group. For each person $i$, his/her temporal pose sequence is first processed by a person-level LSTM to obtain its representation $\bm{h}_i^{t} \in \mathbb{R}^{H_p}$ at time $t$.

Group membership is initialized at time step $t=0$ by arbitrarily placing 2 people in 1 group, with the remaining people spread into solo groups. Then for each time step $t-1$, every person decides their group affiliation for next time step $t$ by choosing to join (or stay in) the group inside which people have the strongest interaction with him. The interaction score for two people $i$ and $j$ at $t-1$ is defined as:
\begin{equation} \label{eq:9}
\begin{split}
p_{ij}^{t-1} = \sigma( \bm{W}_{hs}( \bm{W}_{hh}(\bm{h}_{i}^{t-1} + \bm{h}_{j}^{t-1}) + \bm{b}_{hh}) + \bm{b}_{hs}) \\
\end{split}
\end{equation}
The $p_{ij}^{t-1}$ is a scalar score that measures the degree of interaction. $\bm{W}_{hh} \in \mathbb{R}^{H_p \times H_p}, \bm{W}_{hs} \in \mathbb{R}^{H_p \times 1}, \bm{b}_{hh} \in \mathbb{R}^{H_p}, \bm{b}_{hs} \in \mathbb{R}^{1}$ are weights and biases for the state-to-score transformation. $\sigma$ is the sigmoid function. With the interaction score between people, the $i^{th}$ person can then decide his group at $t$, which is denoted by $m^{t}(i)$, as:
\begin{equation} \label{eq:8}
m^{t}(i) = \text{arg}\,\max\limits_{k}\ 
\begin{cases} 
\dfrac{1}{||G_{k}^{t-1}||} \sum \limits_{\substack{j \in G_{k}^{t-1} \\}} p_{ij}^{t-1}, & \ ||G_{k}^{t-1}|| \neq 0 \\
p_{ii}^{t-1}, & \ ||G_{k}^{t-1}|| = 0
\end{cases}
\end{equation}
Note that since argmax is not differentiable, we use softmax with low temperature to approximate it (c.f.\ Gumbel Softmax~\cite{jang2016categoricalGumbel}). 
In short, if the output of $\text{arg}\,\max(\bm{T})$ is represented as a one-hot vector, then $softmax(\bm{T}/\tau)\rightarrow\text{arg}\,\max(\bm{T})$ as $\tau\rightarrow0$.
When measuring the score for the current group of person $i$, $p_{ii}^{t-1}$ will be included to serve as a smoothing term that increases the probability for person $i$ to keep his group unchanged.

Groups also maintain their information via group-level LSTMs as shown by group nodes in Fig.~\ref{fig:overview}(b). After all group memberships are determined for time $t$, each group will update its state by one step of its LSTM:
\begin{equation} \label{eq:10}
\bm{g}_{k}^{t} = \mathbf{LSTM}_{gr}\left(\dfrac{1}{||G_{k}^{t-1}||} \sum \limits_{\substack{i \in G_{k}^{t-1}}} \bm{W}_{hg}\bm{h}_{i}^{t-1};\bm{g}_{k}^{t-1}\right)
\end{equation}
where $\bm{g}_{k}^{t} \in \mathbb{R} ^{G}$ is the state of $k^{th}$ group at $t$ and $\mathbf{LSTM}_{gr}$ is the group-level LSTM cell. Projection weights $\bm{W}_{hg} \in \mathbb{R}^{H_p \times G}$ project person states to the space of group states.
%
The group state $\bm{g}_{k}^{t}$ will then serve as the interaction context of the $i^{th}$ target at time step $t$ for all $i \in G_{k}^{t}$.

In summary, we use the input pose sequences to predict which groups people in the scene belong to.  Each person has affinity for people with related pose sequences.  Each group has a feature representation based on the people who have been in the group over time.  We use these group feature representations as interaction context for our pose prediction tasks.

\subsubsection{Hierarchical Pose Prediction}
\label{sec:multigran-layer-2}
We generate future poses using the predicted group memberships and encodings of observed pose sequences.  The generation is a hierarchical process in a recurrent neural network framework.  We process a given input sequence of poses from time 1 to $T_1$ and produce an output of predicted poses from time $T_1+1$ to $T_1+T_2$.

The recurrent network takes as input the encoding of the observed poses $\bm{h}_i^{T_1}$, and the group state $\bm{g}_{m^{T_1+1}(i)}^{T_1+1}$ for person $i$.
It generates $\bm{h}_i^{T_1+1}$ by forwarding one step. This new encoding is used to generate the next pose $\widetilde{\bm{P}}_i^{T_1+1}$. The group states are then updated using all new encodings to obtain the $\bm{g}_{m^{T_1+2}(i)}^{T_1+2}$ values. The process is repeated to generate all predicted poses until time $T_1+T_2$.





To allow for finer adjustment of each pose joint, the second layer of our hierarchical model, \textit{refiner LSTM}, takes spatial relations among joints into consideration using spatio-temporal LSTM~\cite{liu2016st-LSTM}. With first-layer prediction as its extra input, the fine-granularity LSTM produces refinement vectors for joints based on (1) the states of the current joint at the previous time step and (2) the states of the previous joint at the current time step. The spatial order is defined based on the kinematic tree.
This produces the final generated sequence of poses $\hat{\bm{P}}_i^{t}$.

\subsubsection{Two-stage Training and Loss Function}
\label{sec:multigran-train}
The multi-granularity LSTM is trained with a two-stage scheme. In the first stage, only the person-level LSTM with the interaction mechanism is trained to produce a reliable first-phase output. The loss is:
\begin{equation}
        \mathcal{L}_{stage \mhyphen 1} = \mathcal{L}_{MSE \mhyphen 1} = \frac{1}{N}\sum_{i=1}^N\sum_{t=T_1+1}^{T_1+T_2}||\bm{\widetilde{P}}_i^t - \bm{P}_i^t||^2
\end{equation}
where $\mathcal{L}_{MSE \mhyphen 1}$ is the pose MSE loss of the first-layer model. 

After finishing the first-stage training, we train the whole model altogether with loss
\begin{equation}
    \label{mg-loss}
        \mathcal{L}_{stage \mhyphen 2} = \mathcal{L}_{MSE \mhyphen 2} + w_{s_1}*\mathcal{L}_{MSE \mhyphen 1}
\end{equation}
where $L_{MSE \mhyphen 2}$ is the pose MSE loss of model's final output defined by
\begin{equation}
    \mathcal{L}_{MSE \mhyphen 2} = \frac{1}{N}\sum_{i=1}^N\sum_{t=T_1+1}^{T_1+T_2}||\bm{\hat{P}}_i^t - \bm{P}_i^t||^2
\end{equation}


\subsection{Adaptive Rendering Network}
\label{sec:adaptive_render}
After getting the pose predictions for each target from the first part of the architecture, the next step of our model is to synthesize for each target a realistic image of the target in the predicted pose. We represent the pose of every person using a \textit{posemap} image in which white body joint points are drawn on a black background canvas.  To accomplish this goal, we propose an adaptive rendering structure where the appearance filters are adaptively computed from an input reference image using a fully convolutional neural network (FCN).  By incorporating this FCN into an encoder-decoder network a realistic image of a target consistent with the desired action and appearance can be generated.


\subsubsection{Network Structure}
 
Fig.~\ref{fig:overview}(c) shows our adaptive rendering network (Ada-R Network) architecture, which consists of two branches: an encoder-decoder branch, and an adaptive rendering branch. The network requires two input images: a posemap image, and a reference image which provides the appearance of the same person in a previous frame. The goal of the network is to generate a realistic image of a person consistent with posemap and having appearance consistent with the reference image.
 
\textbf{Encoder-Decoder}: Instead of training an encoder-decoder network which can reconstruct input images, our encoder-decoder branch shown in Fig.~\ref{fig:overview} is a sketch $\rightarrow$ image model. 

We use the same input size and encoder-decoder structure as in~\cite{isola2016pix2pix}: both generator and discriminator use modules of the form convolution-BatchNorm-Relu~\cite{ioffe2015batch}, the encoder consists of 8 convolutional layers with stride 2 and symmetrically the decoder consists of convolutional layers with fractional stride $\frac{1}{2}$. We use filters of size $5 \times 5$. We also explore a more compact encoder-decoder network by reducing the number of convolutional and fractional strided convolutional layers in our encoder and decoder to 5.

\textbf{Adaptive Rendering}: The encoder-decoder network takes binary posemap images as inputs which do not contain any information about the uniform or clothes of the person.  Hence, we propose to use another network to learn appearance information. By combining these two networks together we are able to generate realistic images of a person wearing the desired clothing. Here we introduce our Ada-R network. 

To transfer the desired appearance to the encoder-decoder branch, we replace the last convolutional filter in the encoder-decoder branch with our adaptive appearance transfer filter. The adaptive appearance filter $K_{ada \mhyphen app}$ encoding appearance information of a person is derived from an input appearance reference image $\bm{I}_{app}$ using a fully-convolutional network
\begin{equation}
    \bm{K}_{ada \mhyphen app} = FCN(\bm{I}_{app})
\end{equation}
Note the rendering of one person's posemap sequence only requires one reference image, and it can simply be the first input frame for that person. The realistic motion sequence is obtained by performing adaptive appearance rendering frame by frame. The filter is applied to rendering procedure by
\begin{equation}
    \bm{F} = \mathcal{E} (\bm{I}_{pose}) 
\end{equation}
\begin{equation}
    \bar{\bm{F}} = \bm{F} \ast \bm{K}_{ada \mhyphen app}
\end{equation}
where $\mathcal{E}$ is the encoder network, $\bm{I}_{pose}$ is the posemap image and $\ast$ is convolution operation. $F$ is the feature map generated by the encoder network and $\bar{\bm{F}}$ is the feature map after applying the adaptive appearance filter to the feature map $\bm{F}$. The person $\bm{I}_{gen}$ with desired appearance is finally produced with
\begin{equation}
     \bm{I}_{gen} = \mathcal{D} (\bar{\bm{F}})
\end{equation}
where $\mathcal{D}$ is the decoder network.

We propose three types of FCN architectures and all three architectures share same encoder-decoder structure. The first FCN with 5 convolutional layers and outputs filters with size $5 \times 5 \times 10$; the second FCN with 3 convolutional layers and outputs filters with size $5 \times 5 \times 10$; the third FCN with 3 convolutional layers and outputs filters with size $5 \times 5 \times 56$.

\subsubsection{Loss Function}

Our network is trained in an adversarial setting, where the Ada-R network is the generator $G$, and a discriminator $D$ is introduced to discriminate between the real and generated images. Let $\bm{I}_{goal}$ be the target image that we try to produce, and $\bm{I}_{gen}$ be the image that Ada-R network generated. The loss $\mathcal{L}$ of Ada-R network is defined as
\begin{equation}
\mathcal{L_{CGAN}}(G, D) + \mathcal{L_{T}}
\end{equation}



Where the appearance transfer loss $\mathcal{L_{T}}$ is defined as
\begin{equation}
\alpha \mathcal{L_{MSE}}(\bm{I}_{gen}, \bm{I}_{goal}) + \beta \mathcal{L_{C}}(\bm{I}_{gen}, \bm{I}_{goal}) + \gamma \mathcal{L_{S}}(\bm{I}_{gen}, \bm{I}_{app}).
\end{equation}
$\mathcal{L_{MSE}}$ is the pixel level MSE loss between generated image and the target image, which is defined as
\begin{equation}
\mathcal{L_{MSE}}(\bm{I}_{gen}, \bm{I}_{goal}) = ||\bm{I}_{gen}-\bm{I}_{goal}||^2.
\end{equation}
$\mathcal{L_{C}}$ and $\mathcal{L_{S}}$ are the content and style loss defined the same as Gatys et al.~\cite{gatys2015neuralStyleTransfer}
\begin{eqnarray}
\mathcal{L_{C}}(\bm{I}_{gen}, \bm{I}_{goal}) = \sum_{l \in {l_{c}}}||F_{l}(\bm{I}_{gen})-F_{l}(\bm{I}_{goal})||^2 \\
\mathcal{L_{S}}(\bm{I}_{gen}, \bm{I}_{app}) = \sum_{l \in {l_{s}}}||G_{l}(\bm{I}_{gen})-G_{l}(\bm{I}_{app})||^2
\end{eqnarray}
where $F_{l}$ is the feature map from layer $l$ of a pretrained VGG-19 network~\cite{Simonyan15VGG}. $l_{c}$ are layers of VGG-19 used to compute the content loss. $G_{l}(\cdot)$ is the Gram matrix which learns the correlations of color distribution given two input images. $l_{s}$ are layers of VGG-19 used to compute the style loss.

The final objective is defined as
\begin{equation}
G^{\star} = \arg \underset{G}{\min}\ \underset{D}{\max}\ \mathcal{L_{CGAN}}(G, D) + \mathcal{L_{T}}
\end{equation}

\section{Experiments}
\label{sec:experiments}
We demonstrate our model on the Volleyball dataset ~\cite{Ibrahim_2016_CVPR_volleyball}. We run person detection~\cite{ren2015faster} and tracking~\cite{dlib09} to get tracklets of each player in each clip. Then OpenPose detector~\cite{cao2017realtime} is used to obtain corresponding pose sequences for each tracklet. We follow the data split of original dataset and preprocessing is conducted to filter out instances with less than 10 joints and clips containing less than 10 targets. We get 1262 clips for training and 790 clips for testing. Images of players are cropped and then resized to $256 \times 256$ pixels.  Our model is trained to observe players in 6 input frames and predict their future for the next 5 frames.


\textbf{Training Details}: For the multi-granularity LSTM, the state size of person, group, and joint level LSTM are 256, 256, and 128, respectively. Pose data are normalized to between 0 and 1. We train the model with initial learning rate of \textit{1e-5}. We set $w_{s_1}$ in Eq.~\ref{mg-loss} to 0.1. To prevent gradient explosion in low-temperature softmax, we use the training strategy suggested by Jang et al.~\cite{jang2016categoricalGumbel} and clip gradients as well. To train our Ada-R network, we compute content loss at layer \textit{relu4-2} and style loss at layer \textit{relu1-2}, \textit{relu2-2}, \textit{relu3-2}, \textit{relu4-2} and \textit{relu5-2} of the pre-trained VGG-19 network. We set the learning rate to \textit{1e-3}, $\alpha=5$, $\beta=0.1$, $\sigma$ is set to bring the content and style losses to a similar scale. To make the training stable, in each iteration the generator is updated twice and the discriminator is updated one time.

\subsection{Results of Pose Prediction}
\label{results_pose_prediction}
We compare our multi-granularity LSTM with two baseline models: (1) vanilla LSTM without interaction among targets; (2) model adapted from SocialLSTM~\cite{Alahi2016Social} by replacing the trajectory prediction in the original work with pose and location prediction and use the social pooling as the group interaction mechanism. We also include comparisons among different variants of our model including: (1) MG w/o refine: our multi-granularity model without refinement layer; (2) MG: our whole multi-granularity model.

We evaluate the performance of future pose generation by measuring the distance between the prediction and the exact pose estimation. MSE is a standard metric for this, but is sensitive to localization error. A prediction will have high MSE even if every joint is off by a small number of pixels; in such cases MSE provides limited intuition as to the quality of generation. We define a score to measure whether a joint is correctly predicted within some tolerable range to the exact pose estimation using a piecewise function. Specifically, for each joint of pose estimation $\bm{P}_{i,j}$ we measure how good the prediction $\hat{\bm{P}}_{i,j}$ is by calculating a score
$$
score(\hat{\bm{P}}_{i,j})= \
\begin{cases}
1, & \ ||\bm{P}_{i,j}-\hat{\bm{P}}_{i,j}||_2 < \mu \\ 
e^{\frac{-({||\bm{P}_{i,j}-\hat{\bm{P}}_{i,j}||}_2 - \mu)^2}{2\sigma^2}}, & \ otherwise
\end{cases}
$$
where $||\cdot||_2$ is the $L_2$ norm, $\mu$ and $\sigma$ should be determined according to the size of posemap in a way that high-score prediction is reasonably close to desired target. In our experiments we set $\mu=5$ and $\sigma^{2}=72$: a joint prediction with 5-pixel error in resolution $256 \times 256$ will get full score.

Quantitative measures of our multi-granularity model and the comparisons with baselines are summarized in Tab.~\ref{pose_eval}. The result shows that our multi-granularity LSTM outperforms baselines on predicting future pose. Our one-layer multi-granularity model can generate poses closer to the exact future pose estimation than the model adapted from SocialLSTM, implying our dynamic group-based interaction mechanism is more effective than modeling interactions of nearby people. The refiner layer is able to further improve the prediction result. The comparison with vanilla LSTM shows that considering interactions among targets helps produce better future poses in multi-person scenes.


\begin{table}[h!]
    \centering
    \begin{tabular}{ | c | c | c | c | c | c |}
    \hline
    \makecell{MSE \\ Joint Score} & t=6 & t=7 & t=8 & t=9 & t=10 \\
    \hline
    \makecell{Vanilla\\LSTM}  & \makecell{22.42 \\ 0.466} & \makecell{24.21 \\ 0.444}  & \makecell{26.73 \\ 0.403}  &  \makecell{29.30 \\ 0.365}  & \makecell{31.94 \\ 0.328} \\
    \hline
    \makecell{Social \\ LSTM}  & \makecell{20.45 \\0.505 } & \makecell{25.85 \\0.384 }  & \makecell{29.96 \\0.318}  &  \makecell{33.44 \\0.276} & \makecell{36.23 \\0.248 } \\
    \hline
    \makecell{MG w/o \\ refine} & \makecell{20.06 \\ 0.530} & \makecell{22.64 \\ 0.481} & \makecell{25.53 \\ 0.429} & \makecell{28.32 \\ 0.385} & \makecell{30.94 \\ 0.349}\\
    \hline
    \makecell{MG(ours)} & \makecell{18.92 \\ 0.567} & \makecell{21.82 \\ 0.505} & \makecell{24.84 \\ 0.446} & \makecell{27.75 \\ 0.397} & \makecell{30.40 \\ 0.358} \\
    \hline
    \end{tabular}
    \caption{Different models for future pose prediction.}
    \label{pose_eval}
\end{table}

\subsection{Results of Adaptive Rendering}

We evaluate the generation using two quantitative measures and show qualitative results. We compare our approaches against a baseline of visual analogy making (VAM)~\cite{NIPS2015_5845_analogy} in which, similar to our main model, 8 convolutional or fractional-strided convolutional layers in the encoder and decoder are used, respectively, and are trained using adversarial loss and MSE loss. We also provide comparisons among different architectures (details shown in Tab.~\ref{tbl:different_architecture}) of our model including (1) the \textit{8-5-10} model; (2) the \textit{8-3-56} model; (3) the \textit{8-3-10} model; (4) the \textit{5-5-10} model. To compare different architectures of our Ada-R network, we use posemap images generated from pose estimation results as inputs. To compare different models for pose predictions, we use posemap images from the predicted poses generated by different models as shown in Sec.~\ref{results_pose_prediction} as inputs and use our \textit{8-5-10} model to generate images. Reference images are achieved by cropping players given detection results and resized to $256 \times 256$. Our appearance rendering network generates images of the same size.

\begin{figure*}[!tph]
\center
\includegraphics[width=0.9\textwidth]{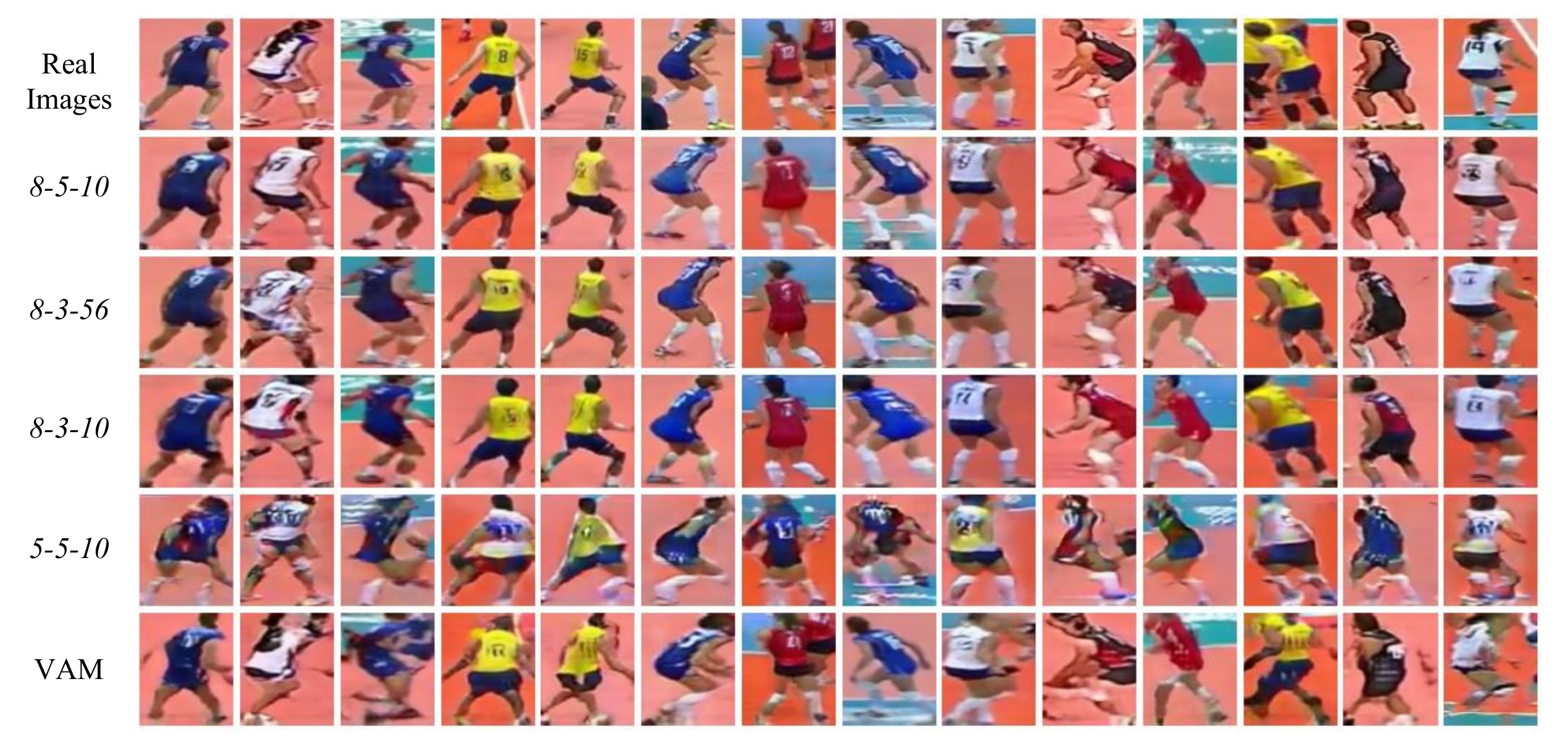}
\caption{Qualitative results of our Ada-R network and the VAM baseline. Given poses from pose estimation, realistic images are generated using the Ada-R network with different architecture settings v.s.\ a VAM baseline.}
\label{fig:ada_gen_part2}
\end{figure*}
\begin{figure*}[!tph]
\center
\includegraphics[width=0.9\textwidth]{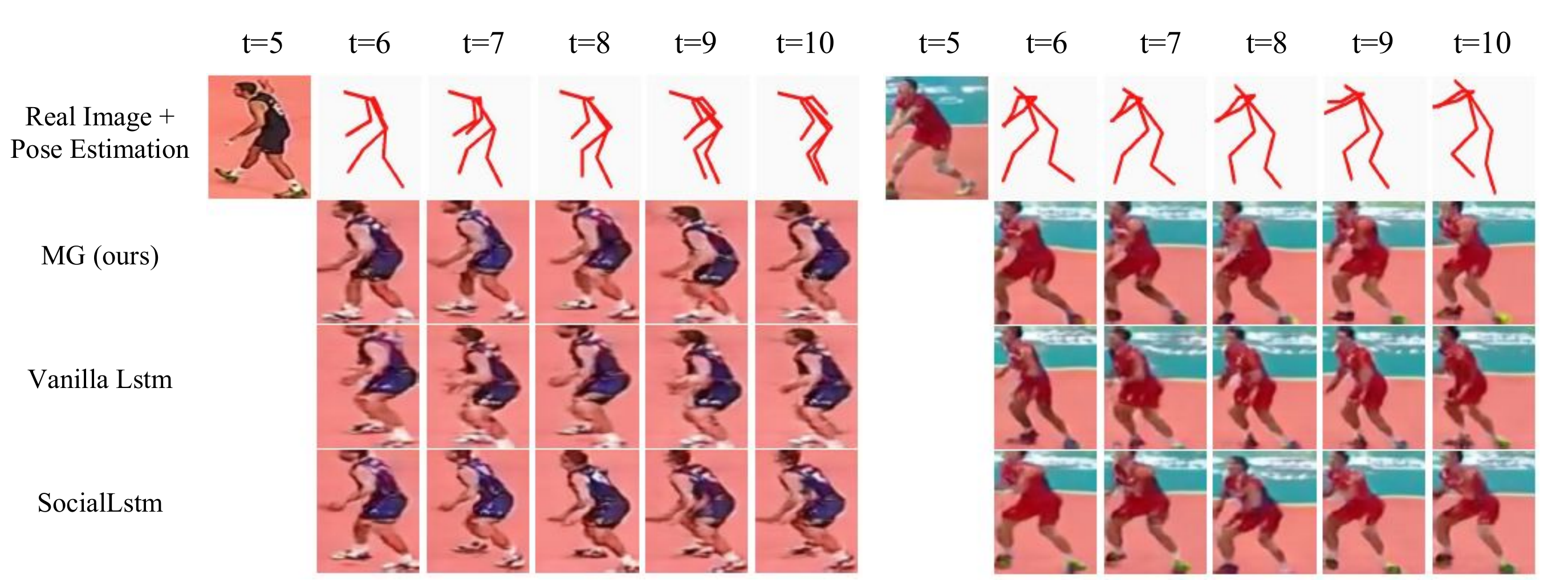}
\caption{Qualitative results of different pose prediction models. Our multi-granularity LSTM can better forecast future poses that are closer to the exact pose estimation results; this results in adaptive rendering results with more realistic, obvious motion. The vanilla LSTM and SocialLstm model cannot generate future pose sequences with obvious changes over time.}
\label{fig:ada_gen_part1}
\end{figure*}

We adopt two evaluation metrics: (1) action classification over sequence; (2) MSE error and PSNR over sequence. An action classifier is trained using real video sequences over the 9 action classes in the training set and tested by using sequences generated by different models in the test set. Since the actions in this dataset are highly unbalanced, we report action classification accuracy on the overall dataset, and accuracy excluding the majority action \textit{standing}. Quantitative measures are shown in Tab.~\ref{tbl:action_classification} and Tab.~\ref{tbl:mse_psnr}. Visualizations are provided in Fig.~\ref{fig:ada_gen_part2} and Fig.~\ref{fig:ada_gen_part1}.

\begin{table}[h!]
  \centering
  \begin{tabular}{ | c | c | c | c | c |}
    \hline
    & \textit{8-5-10} & \textit{8-3-56} & \textit{8-3-10} & \textit{5-5-10} \\ \hline
    layer \# in E-D & 8 & 8 & 8 & 5  \\ \hline
    conv layer \# in FCN & 5 & 3 & 3 & 5  \\ \hline
    ada-filters \# learnt & 10 & 56 & 10 & 10 \\ \hline
  \end{tabular}
  \caption{Different architectures of the Ada-R network.}
  \label{tbl:different_architecture}
\end{table}

\newcommand{\NA}{---}
\begin{table}[h!]
  \centering
  \begin{tabular}{ | c | c | c | c | c | c |}
    \hline
    & \makecell{Real \\ Images} & \makecell{MG\\(ours)} & \makecell{MG w/o \\ refine} & \makecell{Vanilla\\LSTM} & \makecell{Social \\ LSTM}
    \\ \hline
    \makecell{overall \\ non-stand} & \makecell{0.784 \\ 0.440} & \makecell{ 0.672 \\ 0.315} & \makecell{ 0.677 \\ 0.320} & \makecell{ 0.662 \\ 0.278} & \makecell{ 0.655 \\ 0.299}
    \\ \hline
    & \textit{8-5-10} & \textit{8-3-56} & \textit{8-3-10} & \textit{5-5-10} & VAM
    \\ \hline
    \makecell{overall \\ non-stand} & \makecell{ 0.648 \\ 0.339} & \makecell{ 0.656 \\ 0.289} & \makecell{ 0.652 \\ 0.265} & \makecell{ 0.542 \\ 0.229} & \makecell{ 0.668 \\ 0.185}
    \\ \hline
    \end{tabular}
  \caption{Action classification accuracy on generated sequence.}\label{tbl:action_classification}
\end{table}

\begin{table}[h!]
  \centering
  \begin{tabular}{ | c | c | c | c | c | c |}
    \hline
    & \makecell{Real \\ Images} & \makecell{MG\\(ours)} & \makecell{MG w/o \\ refine} & \makecell{Vanilla\\LSTM} & \makecell{Social \\ LSTM}
    \\ \hline
    \makecell{MSE \\ PSNR} & \NA & \makecell{ 0.884 \\ 68.784} & \makecell{ 0.892 \\ 68.575} & \makecell{ 0.901 \\ 68.343} & \makecell{ 0.949 \\ 67.215}
    \\ \hline
    & \textit{8-5-10} & \textit{8-3-56} & \textit{8-3-10} & \textit{5-5-10} & VAM
    \\ \hline
    \makecell{MSE \\ PSNR} & \makecell{ 0.835 \\ 70.095} & \makecell{ 0.906 \\ 68.359} & \makecell{ 0.979 \\ 66.615} & \makecell{ 1.103 \\ 63.564} & \makecell{ 1.045 \\ 65.087}
    \\ \hline
    \end{tabular}
  \caption{MSE and PSNR over generated sequence.}\label{tbl:mse_psnr}
\end{table}

The experimental results in Tab.~\ref{tbl:action_classification} suggest that our Ada-R model can generate realistic sequences with more obvious motions while visual analogy making cannot capture the finer changes in poses and generate sequences with stable motions. Our multi-granularity LSTM can better forecast future poses of players compared with the two baselines: vanilla LSTM and SocialLSTM. Tab.~\ref{tbl:mse_psnr} suggests that the quality of generated images of our Ada-R model is better and is more similar to the generation target compared with visual analogy making. The decreases of MSE and the increases of PSNR over vanilla LSTM and SocialLSTM suggest that our model can better forecast future poses which can benefit the adaptive rendering. Both tables suggest the \textit{8-5-10} model can produce images with better quality, more obvious motion, and achieves the best performance.

Fig.~\ref{fig:ada_gen_part2} shows that most of our Ada-R architectures can generate more realistic images with both action and appearance consistent with the target images, while visual analogy making can generate images with correct appearance but distorted pose, implying explicitly encoding appearance information with filters learned from extra reference images can better disentangle the appearance and pose representations. Fig.~\ref{fig:ada_gen_part1} shows how our \textit{8-5-10} model generates images given different pose prediction results. It is clear that our proposed model can better forecast future pose sequences with obvious motion more similar to the generation targets.

\begin{figure}[h!]
\center
\includegraphics[scale=0.4]{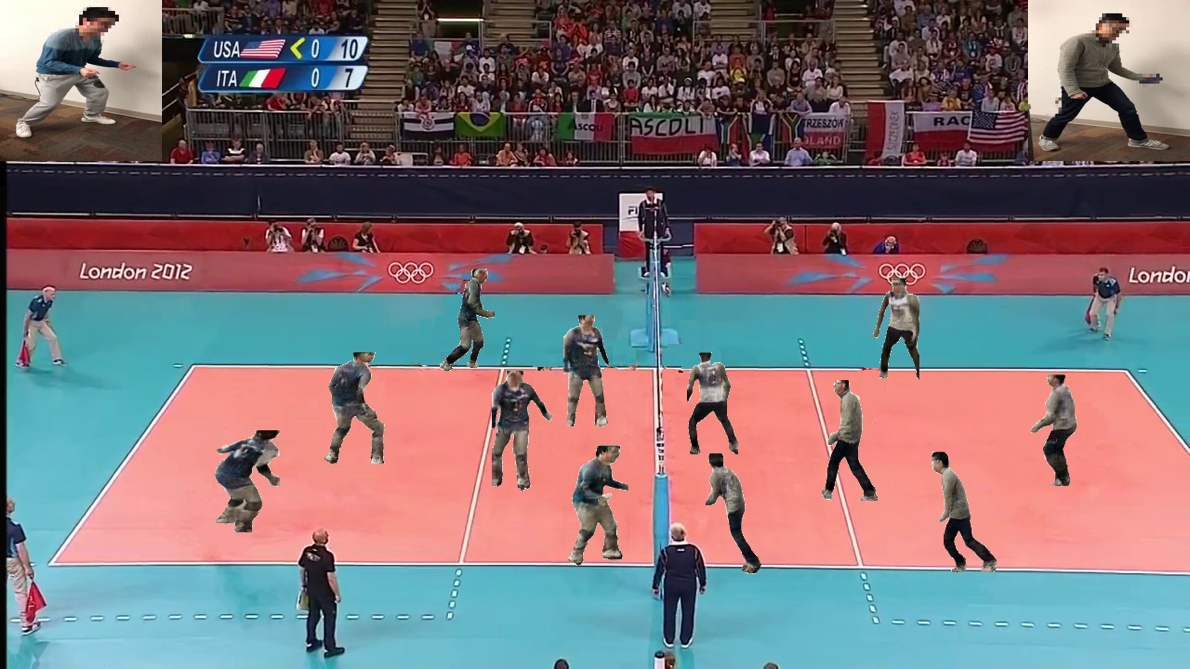}
\caption{Hallucinating people in a volleyball game.}
\label{hallucinating_frame}
\end{figure}

\subsection{Hallucinating People in a Volleyball Game}

Given a set of generated realistic images of two people obtained by fine-tuning models trained on volleyball dataset for extra iterations on the videos of the two people and a background image of volleyball court which is obtained by inpainting the players in a raw frame of resolution $1280 \times 720$, we hallucinate people in a volleyball game (shown in Fig.~\ref{hallucinating_frame}) by segmenting the people out of generated realistic images and copy them to the background image. The two real images of the two people in the top left and right corners are the reference images we use for adaptive rendering.

\section{Conclusion}

We proposed a novel approach for forecasting complex human activity videos. The proposed approach first forecasts future poses using a hierarchical temporal model and then generates realistic images corresponding to the pose by adaptively rendering the appearance from a reference image. Both quantitative and qualitative results show that our model is superior to state-of-the-art approaches and can generate better predictions involving complex human activities. The success of our model demonstrates that the proposed dynamic group-based interaction mechanism can benefit analysis of complex human activity in videos and provide high quality intermediate representations for later image-to-image translation. The proposed adaptive rendering network can render the desired target appearance while adapting to the predicted pose.

\newpage
{\small
\bibliographystyle{ieee}
\bibliography{egbib}
}

\end{document}